\documentclass{article}

    \usepackage[preprint]{neurips_2025}

\usepackage[utf8]{inputenc} % allow utf-8 input
\usepackage[T1]{fontenc}    % use 8-bit T1 fonts
\usepackage{hyperref}       % hyperlinks
\usepackage{url}            % simple URL typesetting
\usepackage{booktabs}       % professional-quality tables
\usepackage{amsfonts}       % blackboard math symbols
\usepackage{nicefrac}       % compact symbols for 1/2, etc.
\usepackage{microtype}      % microtypography
\usepackage{xcolor}         % colors

\usepackage{graphicx}
\usepackage{amsmath}
\usepackage{multirow}
\usepackage{pifont}
\usepackage{colortbl}
\usepackage{subcaption}
\definecolor{tabfirst}{rgb}{1, 0.7, 0.7} % red
\definecolor{tabsecond}{rgb}{1, 0.85, 0.7} % orange
\definecolor{tabthird}{rgb}{1, 1, 0.7} % yellow

\title{Drive Any Mesh: 4D Latent Diffusion for Mesh Deformation from Video}

\author{
Yahao Shi\textsuperscript{\rm 1} 
~~~Yang Liu\textsuperscript{\rm 2} 
~~~Yanmin Wu\textsuperscript{\rm 3} 
~~~Xing Liu\textsuperscript{\rm 4}
~~~Chen Zhao\textsuperscript{\rm 4} 
~~~Jie Luo\textsuperscript{\rm 2} 
~~~Bin Zhou\textsuperscript{\rm 1} \thanks{Corresponding authors.}
\vspace{5pt}
\\
$^1$State Key Laboratory of Virtual Reality Technology and Systems, Beihang University \\
$^2$State Key Laboratory of Software Development Environment
Beihang University \\
$^3$ School of Electronic and Computer Engineering, Peking University \qquad $^4$Baidu VIS    
\vspace{2pt}
}

\begin{document}

\maketitle

\begin{abstract}
We propose \textbf{DriveAnyMesh}, a method for driving mesh guided by monocular video. Current 4D generation techniques encounter challenges with modern rendering engines. Implicit methods have low rendering efficiency and are unfriendly to rasterization-based engines, while skeletal methods demand significant manual effort and lack cross-category generalization.  Animating existing 3D assets, instead of creating 4D assets from scratch, demands a deep understanding of the input's 3D structure. To tackle these challenges, we present a 4D diffusion model that denoises sequences of latent sets, which are then decoded to produce mesh animations from point cloud trajectory sequences. These latent sets leverage a transformer-based variational autoencoder, simultaneously capturing 3D shape and motion information. By employing a spatiotemporal, transformer-based diffusion model, information is exchanged across multiple latent frames, enhancing the efficiency and generalization of the generated results. Our experimental results demonstrate that \textbf{DriveAnyMesh} can rapidly produce high-quality animations for complex motions and is compatible with modern rendering engines. This method holds potential for applications in both the gaming and filming industries.
\end{abstract}

\section{Introduction}
\label{sec:intro}
With the rapid development of 3D generative techniques, creating static 3D assets through prompts such as text or images has become increasingly feasible~\cite{3dsurvey}. However, these static 3D assets are limited in accurately capturing the real world’s dynamic nature. This underscores the importance of 4D generation~\cite{makeavideo3d}, which aims to produce temporally consistent motion for 3D objects while preserving their geometric shape and visual appearance. High-quality 4D generation can greatly enhance the realism of digital environments, with potential applications in fields such as video games, animated films, and virtual and augmented reality (VR/AR).

Recent advances in 4D generation~\cite{makeavideo3d, 4dfy, dreamgaussian4d, consistent4d, stag4d, dreammesh4d}, often leverage pre-trained generative models, such as multi-view models~\cite{zero123} or video generation models~\cite{sv3d}, as well as techniques like score distillation sampling (SDS)~\cite{dreamfusion}. These methods generate 4D content from scratch, but their reliance on time-consuming SDS optimization typically results in long generation times, often taking several hours to produce dynamic outputs. Additionally, such approaches suffer from challenges related to multi-view consistency and are frequently difficult to integrate with modern rendering engines. While existing methods focus on generating 4D assets from scratch, a more practical approach may involve leveraging the large pool of already available 3D assets. These can be either handcrafted or procedurally generated, offering a more efficient means of 4D content creation. One such approach is to bind a skeleton to a 3D model, using it to drive the motion of the object~\cite{magicpose4d, mimo}. However, the time-consuming task of designing skeletons for each 3D asset presents significant challenges, limiting the scalability and generalizability of this method.

In this paper, we propose \textbf{DriveAnyMesh}, a novel framework for 4D asset generation that overcomes the limitations of previous approaches. By utilizing video-guided mesh deformation, we drive existing 3D assets, eliminating the need for manual skeleton binding. Inspired by modern rendering engine pipelines, we treat motion as a sequence of point cloud trajectories. Given a 3D model and a monocular video, we generate motion trajectories for each vertex of the model. These vertex trajectories define the motion of the 4D asset. Our method bypasses the need for skeleton binding, offering a more generalizable solution to 4D asset generation.

We create a dataset of 40,000 textured and 10,000 untextured models with motion to train our model. For each model, we render multi-view images of the animation and save the vertex coordinates for each animation frame. Building upon the work of 3DShape2VecSets~\cite{3dshape2vecset}, we adopt latent sets as our 4D representation. Our approach employs a latent diffusion model, where a transformer-based variational autoencoder (VAE)~\cite{vae} encodes the geometry of the initial mesh and motion information from multi-view images. We then use diffusion~\cite{DDPM} to denoise latent sets conditioned on the initial mesh and monocular video. The denoised result is passed through the VAE decoder, producing a complete animation as a sequence of point cloud trajectories.

We can generate realistic mesh motion by modifying the mesh vertices according to the obtained point cloud trajectory sequence within modern rendering engines. Our method also enables motion transfer, as the input mesh and video are Disentangled, allowing for flexible adaptation to different assets and motion styles. Experimental results demonstrate that our approach outperforms existing methods~\cite{dreamgaussian4d, stag4d, consistent4d, dreammesh4d} in appearance, geometry, and motion, highlighting its potential for practical applications, such as the rapid conversion of 3D assets into 4D assets.

Our contributions can be summarized as follows: 1) A novel approach to 4D generation that uses monocular video to animate 3D assets; 2) A large-scale 4D asset dataset containing multi-view videos and mesh vertex sequences; 3) A new latent diffusion model-based method with a novel VAE architecture and diffusion structure designed specifically for 4D generation.

\begin{figure*}[t]
  \centering\includegraphics[width=1.0\linewidth]{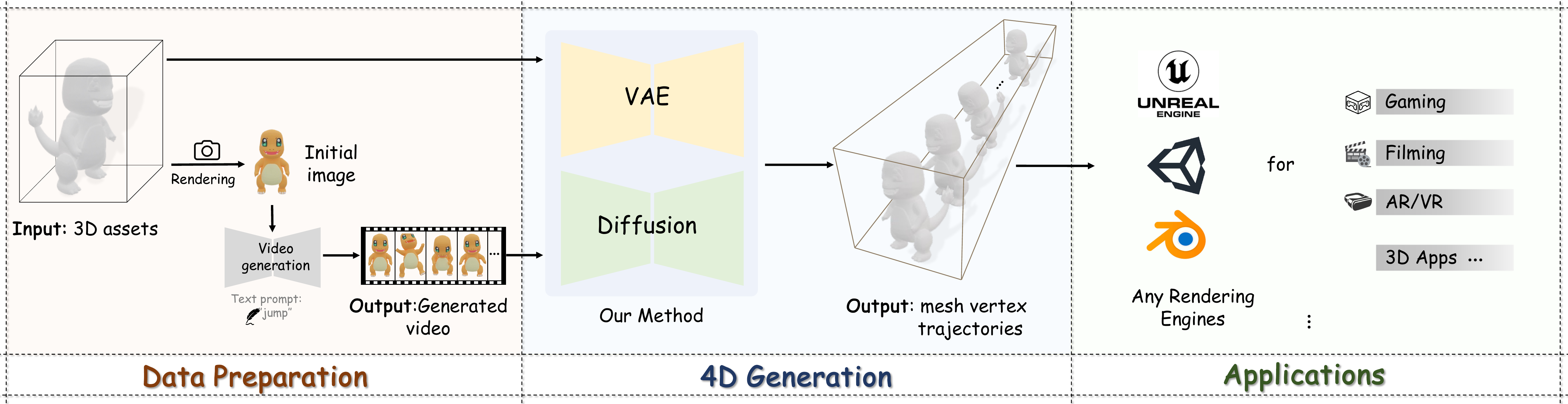}
   \vspace{-18pt}
    \caption{Starting with a 3D asset, \textbf{DriveAnyMesh} leverages monocular videos to produce dynamic animations that seamlessly integrate into modern rendering engines. As shown in the figure, given a 3D asset, we can render a still image and then generate a driving video using video diffusion. Our approach utilize the provided 3D assets and monocular video to generate the motion trajectories for the mesh vertices. Finally, we can drive the 3D asset for applications. }
    \vspace{-12pt}
    \label{fig:teaser}
\end{figure*}

\section{Related Works}
\label{sec:related_works}
\noindent \textbf{Diffusion Models.}
Diffusion models have emerged as a powerful framework for generative tasks, recognized for their capacity to capture complex data distributions through a Markov chain-based denoising process~\cite{DDPM}. Initially successful in text-to-image synthesis~\cite{LatentDiffusion}, these models have been extended to various domains, including video generation and 3D content creation. In video generation, models such as the Video Diffusion Model~\cite{VDM} utilize 3D U-Nets instead of traditional 2D U-Nets to model temporal dynamics effectively. Make-A-Video~\cite{Make-A-Video} extends text-to-image diffusion models to text-to-video generation, eliminating the need for text-video pairs by leveraging temporal layers to ensure frame consistency. AnimateDiff~\cite{Animatediff} introduces a learnable motion module into a pre-trained text-to-image model, maintaining the original model's performance while effectively capturing temporal coherence. In the context of 3D vision tasks, diffusion models have been adapted to manage more intricate structures. For instance, Zero-1-to-3~\cite{zero123} employs a stable diffusion model conditioned on relative camera poses and single images for novel view synthesis. Nonetheless, challenges such as the Janus problem and content drafting persist due to the lack of explicit 3D modeling~\cite{mvdream, wonder3d}. The 3DShape2VecSet~\cite{3dshape2vecset} introduces a novel neural field representation for generative diffusion models, encoding 3D shapes using sets of latent vectors that capture similar local geometric patterns across different objects. This approach uses transformer-based attention mechanisms to enhance 3D shape encoding and generative modeling from surface models or point clouds.

\noindent \textbf{4D Representation.}
Dynamic 3D representations, or 4D representations, are essential for reconstructing and generating dynamic scenes. Existing methods often adapt static Neural Radiance Fields (NeRF) to dynamic scenarios, resulting in deformable NeRFs~\cite{dnerf, nerfies} and time-varying NeRFs~\cite{hexplane, kplane}. While these implicit representations can model complex dynamics, they suffer from long optimization times and lower reconstruction quality due to the computational demands of volume rendering. To address these limitations, recent research has shifted towards explicit or hybrid representations that offer faster rendering speeds and enhanced quality. The 3D Gaussian Splatting~\cite{3dgs} method introduces anisotropic 3D Gaussian representation and tile-based differentiable rasterization, significantly advancing due to its explicit nature and real-time rendering capabilities. Some existing works focus on training networks to predict Gaussian kernel deformations and optimize per-frame outputs\cite{dynamic3dgaussian}. Additionally, SC-GS~\cite{SCGS} and HiFi4G~\cite{hifi4g} employ sparse control points for Gaussian kernel deformation, with SC-GS utilizing Linear Blend Skinning (LBS) and HiFi4G applying Dual Quaternion Skinning (DQS) to ensure spatiotemporal consistency. DreamMesh4D~\cite{dreammesh4d} implements a hybrid approach that combines LBS and DQS, alleviating the drawbacks of both methods. Our work employs latent sets to encode geometry and motion, allowing these sets to query deformation using explicit points. This 4D representation is sufficiently flexible to capture complex motion patterns, generalizes to unseen motions, and is well-suited for integration with latent diffusion models.

\noindent \textbf{4D Generation.}
Extending 3D generation into the spatiotemporal domain, 4D generation aims to synthesize dynamic 3D content from various inputs such as text prompts, single images, monocular videos, or static 3D assets. This task necessitates consistent geometry prediction and the generation of temporally coherent dynamics. Early efforts, such as MAV3D~\cite{makeavideo3d}, utilize score distillation sampling (SDS) derived from video diffusion models to optimize dynamic NeRF representations based on textual prompts. Subsequent research, including 4D-FY~\cite{4dfy}, combines supervisory signals from image, video, and 3D-aware diffusion models to enhance the structure and appearance of generated 4D models. Recent studies have also investigated 4D generation from monocular videos. Consistent4D~\cite{consistent4d} introduces an interpolation-driven loss between adjacent frames to improve spatiotemporal consistency, although it lacks cross-frame temporal modeling. STAG4D~\cite{stag4d} employs a multi-view diffusion model to initialize multi-view images based on input video frames and implements a fusion strategy to enhance temporal consistency. Despite these advancements, many of these methods encounter challenges such as slow optimization speeds and color oversaturation due to the application of SDS loss~\cite{dreamfusion}. Diffusion4D~\cite{diffusion4d} and SV4D~\cite{sv4d} have developed 4D-aware video diffusion models capable of synthesizing orbital views of dynamic 3D assets by leveraging a carefully curated dynamic 3D dataset. However, generating multi-view consistent videos remains challenging due to these approaches' absence of explicit 3D modeling.

\section{Method}
\label{sec:method}

\subsection{Motivation and Objective}
\label{sec:motivate}
Compared to static 3D assets, generating 4D dynamic assets is significantly more challenging, especially when considering compatibility with modern rendering engines. Existing methods often create dynamic models from scratch, relying on Score Distillation Sampling (SDS), which results in long processing times and issues like oversaturated colors. Additionally, the output meshes typically lack the necessary correspondences between frames, making them insufficient for practical applications. Another approach is adding skeletal structures to these models, which requires substantial effort to design and bind the skeletons. To address these challenges, we propose a video-driven method for generating dynamic animations from static 3D assets. This method outputs the vertex positions of the 3D mesh for each keyframe in a video, allowing us to bypass skeletal binding and directly achieve mesh deformations. The resulting animations can be integrated into modern rendering engines.

Formally, given an initial point cloud, $P_1 = \{p_i \in \mathbb{R}^3\}^N_{i=1}$, sampled from a mesh and a monocular video, $\mathcal{V} = \{I_t \in \mathbb{R}^{H\times W\times 3}\}^T_{t=1}$, captured using a fixed camera pose, we aim to output a sequence of point clouds, $\mathcal{P} = \{P_t \in \mathbb{R}^{N\times 3}\}^T_{t=1}$. Each point cloud contains $N$ points, and the video comprises $T$ images captured at timestamps, $t\in\{1,\dots,T\}$, where $H$ and $W$ are the height and width of the images, respectively. Therefore, we propose a 4D latent diffusion model that can generate denoised samples from the learned conditional distribution, $p(\mathcal{P}|P_1, \mathcal{V})$. Consequently, we obtain an animation formed from the sequence of point clouds.

\begin{figure*}[t]
  \centering   \includegraphics[width=1.0\linewidth]{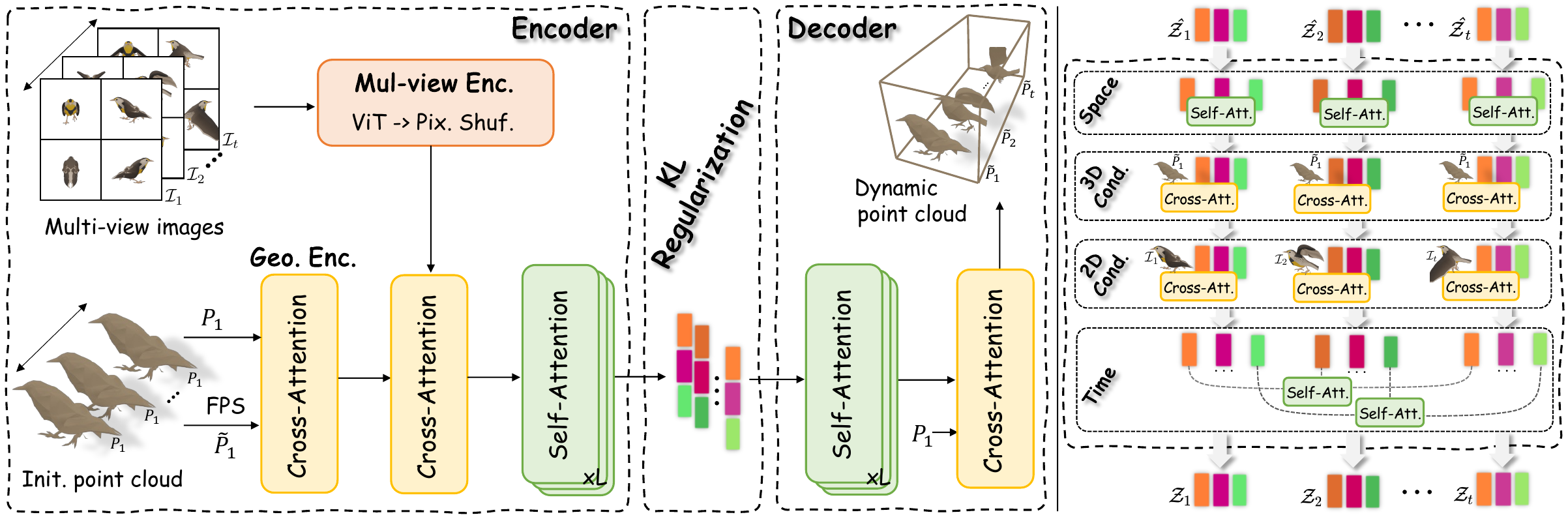}
  \vspace{-18pt}
   \caption{Overview of the proposed \textbf{DriveAnyMesh} framework. It is based on 4D latent diffusion, where we utilize latent sets as a motion representation that encodes both geometry and deformation between two frames (\textit{e.g.}, ($P_1$, $\mathcal{I}_1$), ($P_1$, $\mathcal{I}_2$), $\dots$, ($P_1$, $\mathcal{I}_t$)). On the left, we employ a transformer-based VAE to learn the motion representation conditioned on multi-view images. This learned latent set can then be used to deform the initial point cloud. On the right, we capture spatiotemporal information using a transformer-based diffusion model conditioned on both the initial point cloud and a monocular video. Further details are provided in Sec.~\ref{sec:method}.}
   \label{fig:pipeline}
   \vspace{-12pt}
\end{figure*}

\subsection{Data Preparation}
\label{sec:data_preparation}
The development of advanced 4D diffusion models requires access to high-quality 4D assets. In response to the scarcity of large-scale training datasets featuring dynamic 3D objects, we have curated a new 4D dataset derived from the expansive Objaverse dataset~\cite{objaverse, objaversexl}, which contains a significant number of annotated 3D objects. Given that the original Objaverse dataset primarily comprises static 3D assets, many of which are of low quality (\textit{e.g.}, partial scans, missing textures), we implemented a series of empirical rules to filter the dataset effectively. The curation process includes the following steps: 1) \textbf{Model Selection.} We focus on 3D assets labeled as “Animated” within the Objaverse datasets. 2) \textbf{Data Curation.} We render images of these 3D assets to curate our dataset. A set of filtering criteria was established to select models that clearly represent distinct objects, such as people, animals, and machinery while ensuring that their color schemes and movements adhere to conventional expectations. Specifically, we exclude scene models, geometrically anomalous objects (including partial scans, geometrically damaged items, and distorted surfaces), and models exhibiting atypical movements (such as abstract motion, animated playback, or fluid dynamics like flowing water or smoke). Finally, we label the models to indicate whether they are classified as “textureless models.” 3) \textbf{Geometric and Appearance Filtering.} For the remaining 4D assets, we render T multi-view images according to predefined camera poses under random environment lights and save the mesh's point cloud sequence. Additionally, we address instances of minimal movement by calculating the Structural Similarity Index (SSIM) using the rendered images and assess whether the point cloud exceeds designated boundaries to manage cases where assets extend beyond scene limits.

\noindent
This comprehensive strategy ensures the inclusion of only suitably positioned dynamic assets, resulting in a high-quality collection of 4D objects. Finally, we compile 40,000 high-quality animated assets, along with 10,000 animated assets categorized as textureless models. These dynamic 3D assets will be made available as part of our curated dataset. The curated 4D dataset is a valuable resource for training and evaluating 4D diffusion models, enhancing capabilities for dynamic content generation in future research.

\subsection{Motion Representation with Latent Sets}
Existing methods typically use deformable NeRF and 3DGS to represent dynamic scenes. However, these approaches are not well-suited for animations in modern rendering engines. In this work, we model dynamic scenes as trajectories of point clouds, where the points correspond to the vertices of 3D assets. The primary challenge lies in the large and variable number of vertices in a mesh, which complicates the encoding of these point clouds within a unified framework. Inspired by 3DShape2VecSets~\cite{3dshape2vecset}, we utilize latent sets, $\mathcal{F} = \{f_i \in \mathbb{R}^C\}^M_{i=1}$, to encode both the geometry and motion of the scene. These learnable latent sets can efficiently query point deformation for all vertices. In the Latent Diffusion model~\cite{LatentDiffusion}, a variational autoencoder (VAE)~\cite{vae} is used to compress images into latent representations. Similarly, to obtain the 4D latent representation, we employ a transformer-based VAE that consists of a 4D motion encoder, a KL regularization block, and a 4D motion decoder, as illustrated in Fig.~\ref{fig:pipeline} (left).

\noindent \textbf{4D Motion Encoder.}
Our Variational Autoencoder encodes mesh geometry and deformation between two keyframes, ($P_1$ and $P_t$), conditioned on multiple images $\mathcal{I}_t = \{I_i\in\mathbb{R}^{H\times W\times 3}\}^v_{i=1}$ at timestamp $t\in\{1,\dots,T\}$, where $T=30$ in our settings. In the geometry head, $\mathrm{Enc}_\mathrm{geo}$, of the encoder, we begin by sampling the mesh of 3D assets using a point cloud of size 
$N=2048$. The sample points are obtained via random sampling (which emphasizes local geometry) and farthest-point sampling (which captures global geometry). Following the approach in 3Dshape2VecSets~\cite{3dshape2vecset}, we subsample the points to a smaller set of $M=512$ points, $\Tilde{P}_1$ = $\{p_i\in\mathbb{R}^3\}^M_{i=1}$, using farthest point sampling. We then apply cross-attention mechanism (denoted as $\mathrm{CroAtt}$) between $\Tilde{P}_1$ and $P_1$ to obtain learnable latent sets after position embedding PEmb: $\mathbb{R}^3 \rightarrow \mathbb{R}^{C}$, as shown in the following equation:

\begin{equation}
    \mathrm{Enc}_\mathrm{geo}(P_1) = \mathrm{CroAtt}(\mathrm{PEmb}(\Tilde{P}_1), \mathrm{PEmb}(P_1)).
\end{equation}

To integrate post-motion geometric information, we introduce a multi-view appearance head, denoted as $\mathrm{Enc}_\mathrm{mul}$, within the encoder. We aim for the learned latent sets to capture geometric features from point clouds and encode information from multi-view image modalities, which provides the foundation for enabling the latent diffusion model to relax its constraints to a monocular video. This multi-view appearance head extracts image features using a Vision Transformer (ViT: $\mathbb{R}^{H\times W\times 9} \rightarrow \mathbb{R}^{H'\times W'\times C}$)~\cite{vit} from images represented with Plücker camera embeddings~\cite{grm}. To improve the resolution of the extracted features, we apply pixel shuffle (Upper: $\mathbb{R}^{H'\times W'\times C} \rightarrow \mathbb{R}^{2H'\times 2W'\times C/2}$)~\cite{pixelshuffle}. Additionally, attention mechanisms are employed to fuse image features from multiple viewpoints, producing the final image representation. This representation encodes appearance features and incorporates geometric features at timestamp $t$, denoted as follows:

\begin{equation}
    \mathrm{Enc}_\mathrm{mul}(\mathcal{I}_t) = \mathrm{Upper}(\mathrm{ViT}(\mathcal{I}_t)), \,\, t \in \{1, \dots ,T\}.
\end{equation}

Given the features of two frames, one obtained from point clouds and the other from images, we use a cross-attention mechanism to fuse the features of these two modalities. This allows each point's feature to query the corresponding changes occurring after the motion, as represented by the image features. The fused features are then processed through $L$ self-attention blocks (denoted as $\mathrm{SelAtt}$), resulting in geometric and motion deformation features at timestamps $1$ to $t$, denoted as follows:

\begin{equation}
    \mathcal{F}_t \leftarrow  \,\, \mathrm{SelAtt}^l(\mathrm{CroAtt}(\mathrm{Enc}_\mathrm{geo}(P_1), \mathrm{Enc}_\mathrm{mul}(\mathcal{I}_t))), 
     \quad t \in \{1, \dots, T\}, \,\, \mathrm{for} \,\, l = 1, \dots, L.
\end{equation}

\noindent \textbf{KL Regularization Block.}
To facilitate the training of the latent diffusion model, we apply a regularization technique to the latent using the Kullback-Leibler (KL) divergence. The latent, denoted as $\mathcal{F}$, are projected linearly to the mean $\mu$ and standard deviation $\sigma$. The compressed latent variables, $\mathcal{Z} = \{z_i\in\mathbb{R}^{C_0}\}^M_{i=1}$ , are then sampled using the reparameterization trick, as expressed by the Equ.~\ref{equ:latent}, where $g_\mu(x): \mathbb{R}^C \rightarrow \mathbb{R}^{C_0}$ $g_\sigma(x): \mathbb{R}^C \rightarrow \mathbb{R}^{C_0}$ are linear layers, and $C_0 = 32$.

\begin{equation}
\label{equ:latent}
    \mu = g_\mu(\mathcal{F}),\,\,\sigma = g_\sigma(\mathcal{F}),\,\,\mathcal{Z} = \mu\mathcal{F} + \sigma*\epsilon,\,\, \epsilon \sim \mathcal{N}(\mathbf{0},\mathbf{1}). 
\end{equation}

The KL regularization can be written as Equ.~\ref{equ:6}. We set the weight for KL regularization as $0.001$.

\begin{equation}
\label{equ:6}
\mathcal{L}_\mathrm{reg}(\mathcal{F}) = \frac{1}{2}(\mu^2+\sigma^2-\log\sigma^2).
\end{equation}

\noindent \textbf{4D Motion Decoder.}
We utilize $L$ self-attention to derive the final features, $\mathcal{F}_t$, as described below:

\begin{equation}
    \mathcal{F}_t \leftarrow \mathrm{SelAtt}^l(\mathcal{F}_t),t \in \{1, \dots, T\}, \mathrm{for} \,\, l = 1, \dots, L.
\end{equation}

Given a point cloud $P_1$ sampled from the initial mesh, we query the deformed point cloud, $\hat{P}_t$, using the features $\mathcal{F}_t$. The optimization is based on a loss function that combines point-level mean squared error (MSE) and Euclidean distance (DIS) as Equ.~\ref{equ:7}, where $\lambda = 0.1$.

\begin{equation}
\label{equ:7}
    \mathcal{L}_\mathrm{def}(\mathcal{F}_t, P_1) = \mathbb{E}_{P_t\in\mathbb{R}^{N\times 3}}[\mathrm{MSE}(\hat{P}_t, P_t)] + \lambda\mathbb{E}_{P_t\in\mathbb{R}^{N\times 3}}[\mathrm{DIS}(\hat{P}_t, P_t)], t\in\{1,\dots, T\}.
\end{equation}

\subsection{Motion Generation with Diffusion Model}
Our transformer-based VAE is designed to encode both geometry and deformation across two frames. To capture the spatiotemporal information across all frames, we train a transformer-based latent diffusion model in the latent space, \textit{i.e.}, the bottleneck in Equ.~\ref{equ:latent}, using the diffusion formulation from EDM~\cite{edm}. Consequently, our denoising objective is given by Equ.~\ref{equ:denoiser}, where $\mathcal{Z}_t$ represents the compressed latent at timestamp $t$ after the KL regularization block, and $P_1$ and $\mathcal{V}$ are the conditions (the mesh and the monocular video), as described in Sec.~\ref{sec:motivate}.

\begin{equation}
\label{equ:denoiser}
    \mathbb{E}_{\epsilon\sim\mathcal{N}(\mathbf{0}, \sigma^2\mathbf{I})}\sum^T_{t=1}\|\mathrm{Denoiser}(\mathcal{Z}_t+\epsilon,\sigma,P_1,\mathcal{V})-\mathcal{Z}_t\|^2_2.
\end{equation}

Given a monocular video $\mathcal{V}$, the denoised latent sequence is denoted as $\{\mathcal{Z}_1, \mathcal{Z}_2,\dots,\mathcal{Z}_t\}^T_{i=1}$, which is encoded by VAE encoder using $P_1$ and $\{\mathcal{I}_1,\mathcal{I}_2,\dots,\mathcal{I}_t\}^T_{i=1}$. In Fig.~\ref{fig:pipeline} (right), our transformer-based denoiser block consists of spatial self-attention, conditional cross-attention and temporal self-attention, as illustrated in Fig.~\ref{fig:pipeline} (right).

We treat the denoised latent $\mathcal{Z}_t\in\mathbb{R}^{M\times C_0}$ as a point cloud of size M. To capture spatial features for each point, we apply self-attention. Then, We add the geometry condition $P_1$ to each denoised latent $\mathcal{Z}_t$ to incorporate condition information. Then, we introduce the appearance condition $\mathcal{V}$ by pairing $\mathcal{Z}_t$ with $I_t$. Finally, we obtain $T$ latent containing both spatial and conditional features, $\{\mathcal{Z}_t\}^T_{t=1}$, which can be treated as point cloud trajectories of size $M$. Temporal features are fused using self-attention across each point's trajectory. \textit{e.g.}, we fuse the features from the first row of each $\mathcal{Z}_t$. All update rules is expressed as Equ.~\ref{equ:10}, where $\epsilon\sim\mathcal{N}(\mathbf{0}, \sigma^2\mathbf{I})$.

\begin{equation}
\label{equ:10}
    \mathcal{Z}_t = \mathrm{CroAtt}(\mathrm{CroAtt}(\mathrm{SelAtt}(\mathcal{Z}_t + \epsilon),P_1), I_t),\,\,\{\mathcal{Z}_t\}^T_{t=1} = \mathrm{SelAtt}(\{\mathcal{Z}_t\}^T_{i=1}).
\end{equation}

\vspace{4pt} 
\noindent
Through $L$ denoiser blocks, we obtain the predicted denoised latent $\hat{\mathcal{Z}}_t$ and train the denoiser using Equ.~\ref{equ:denoiser}. In our implementation, we randomly sample a subset $\mathcal{V'} = \{I_t \in \mathbb{R}^{H\times W\times 3}\}^{T'}_{t=1}$, where $T' = T/3$, during training. This introduces additional variability, as the method does not explicitly use the timestamp $t$ as input.

\subsection{Refinement and Mesh Driven}
Finally, we obtain the mesh driven by the point cloud representation generated from the network output, denoted as $\hat{\mathcal{P}} = \{\hat{P}_t \in \mathbb{R}^{N\times 3}\}^T_{t=1}$. To further refine the output, we introduce a threshold $\delta$. If the Euclidean distance between two points is below this threshold, we assume minimal movement and reset the point to its previous frame position. This step helps mitigate jitter caused by inaccuracies in the network output. Finally, the resulting data is applied to the corresponding vertex, producing the animated meshes.

\begin{figure*}[t]
  \centering
   \includegraphics[width=1.0\linewidth]{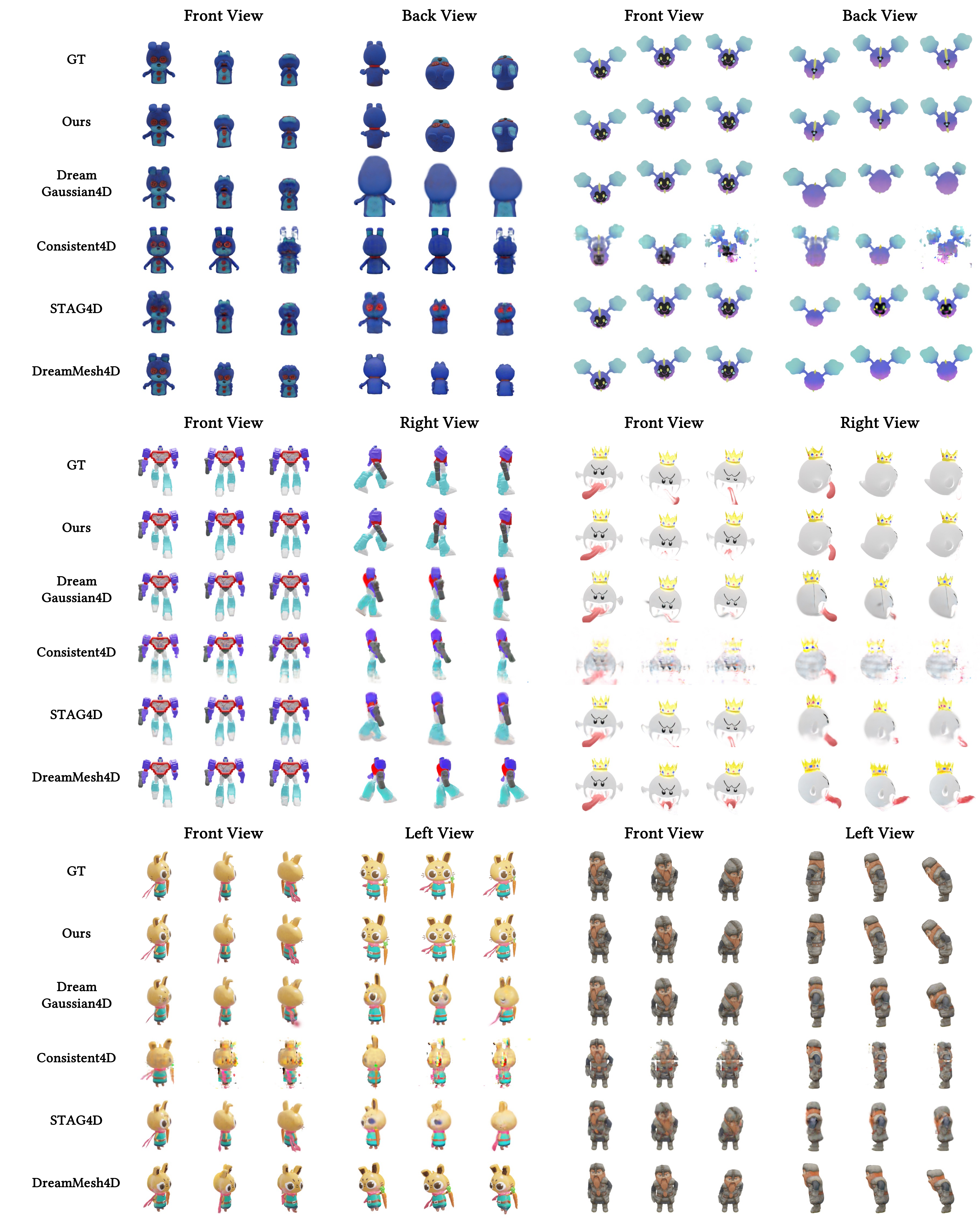}
   \vspace{-18pt}
   \caption{Qualitative comparison of apperance and motion on our proposed datasets.}
   \vspace{-12pt}
   \label{fig:results}
\end{figure*}

\section{Experiment}
\label{sec:experiment}

\subsection{Experiment Setup}
\noindent \textbf{Implementation and Baselines.}
We trained our model on 16 Ascend 910B NPUs. For the VAE, we trained for 200 epochs with a batch size of 1152, which took approximately 90 hours. For the diffusion model, we trained for 200 epochs with a batch size of 512, requiring around 120 hours. We used the Adam optimizer and employed a cosine annealing learning rate schedule. The inference time is approximately 10 seconds for processing 30 frames on a single Ascend 910B NPU. Further details on implementation can be found in the supplementary materials. To ensure reproducibility, we will publicly release the source code. As baseline methods, we used DreamGaussian4D~\cite{dreamgaussian4d}, Consistent4D~\cite{consistent4d}, STAG4D~\cite{stag4d}, and DreamMesh4D~\cite{dreammesh4d}, selected based on their popularity and the availability of their source code.

\noindent \textbf{Metrics and Datasets.} 
To evaluate the appearance quality of all methods, we use Peak Signal-to-Noise Ratio (PSNR), Structural Similarity Index (SSIM), and Learned Perceptual Image Patch Similarity (LPIPS) for each frame. We employ Chamfer Distance for each extracted mesh to assess the geometry quality of all methods. For the qualitative evaluation, we report the Appearance Score (AS) and Motion Score (MS) for each video. As described in Sec.~\ref{sec:data_preparation}, we use 40,000 high-quality animated assets with textures. An asset may have multiple animations. We randomly select 30 key frames for each animation, spaced at intervals no greater than 3 frames. For each frame, we render four images from orthogonal camera viewpoints, each with a $256 \times 256$ resolution. Additionally, we subdivide the mesh to obtain over 50,000 vertices and save the point cloud for each frame. We subsample $10\%$ of all animated assets to form our test set for evaluating the model's performance.

\subsection{Experiment}
\noindent \textbf{Quantitative Experiment.} 
In quantitative experiments, we evaluate the performance of our approach by comparing its appearance and geometric quality with several baseline methods. For the appearance evaluation, we use Blender to render high-quality images from meshes generated by our network’s output. To assess geometric quality, we extract point clouds using 3D Gaussian-based methods~\cite{dreamgaussian4d, stag4d, dreammesh4d} and NeRF-based methods~\cite{consistent4d}. The quantitative results, presented in Tab.~\ref{tab:1}, show that our method outperforms all baselines in both appearance and geometry. Additionally, we provide visual comparisons in Fig.~\ref{fig:results}, clearly illustrating the qualitative improvements achieved by ours. These visual results highlight the advantages of our method in practical applications. Although these methods can generate high-quality results from the frontal view, they still produce poor outcomes from other perspectives due to issues with geometric consistency. When evaluating methods within the SDS framework, we observe that some still struggle with the Janus problem, where inconsistencies in geometry can introduce problematic artifacts. In contrast, our method ensures high-quality appearance by using an existing mesh rather than generating one from scratch, resulting in more consistent and realistic outputs. This represents a significant advantage over methods that rely solely on procedural mesh generation, which often face challenges in maintaining visual consistency.

\begin{table}[t]
\centering
{
\caption{Evaluation results of appearance (\textbf{PSNR}$\uparrow$, \textbf{SSIM}$\uparrow$, \textbf{LPIPS}$\downarrow$), motion (Chamfer Distance$\uparrow$), and User Study (Appearance Score$\uparrow$, Motion Score$\uparrow$) on the proposed dataset.}
\label{tab:1}
\fontsize{9pt}{10.8pt}\selectfont
\begin{tabular}{l|ccc|c|cc}
\toprule
\multirow{2}{*}{Method} & \multicolumn{3}{c|}{App. (4 views)} & \multicolumn{1}{c|}{Geo.}  & \multicolumn{2}{c}{User Study}          \\
                        &  PSNR$\uparrow$ &  SSIM$\uparrow$ &  LPIPS$\downarrow$  & CD$\downarrow$  & AS$\uparrow$   & MS$\uparrow$ \\ 
\midrule

DreamGaussian4D~\cite{dreamgaussian4d} & 15.842 & 0.839 & \cellcolor{tabthird}0.139 &  
\cellcolor{tabthird}0.057 & \cellcolor{tabthird}3.01 & \cellcolor{tabthird}2.94 \\
Consistent4D~\cite{consistent4d} & \cellcolor{tabthird}16.583 & \cellcolor{tabthird}0.847 & 0.175 &  
0.221 & 1.05 & 1.08 \\
STAG4D~\cite{stag4d} & \cellcolor{tabsecond}17.034 & \cellcolor{tabsecond}0.855 & \cellcolor{tabsecond}0.131 &  
0.086 & 2.66 & 2.61 \\
DreamMesh4D~\cite{dreammesh4d} & 15.902 & 0.838 & 0.140 &  
\cellcolor{tabsecond}0.048 & \cellcolor{tabsecond}3.33 & \cellcolor{tabsecond}3.42 \\
\midrule
Ours & \cellcolor{tabfirst}24.391 & \cellcolor{tabfirst}0.950 & \cellcolor{tabfirst}0.030 & \cellcolor{tabfirst}0.018 & \cellcolor{tabfirst}4.91 & \cellcolor{tabfirst}4.92 \\ 
\bottomrule
\end{tabular}}
\vspace{-12pt}
\end{table}

\begin{figure}[t]
    \centering
    \begin{subfigure}[t]{0.48\linewidth}
        \centering
        \includegraphics[width=\linewidth]{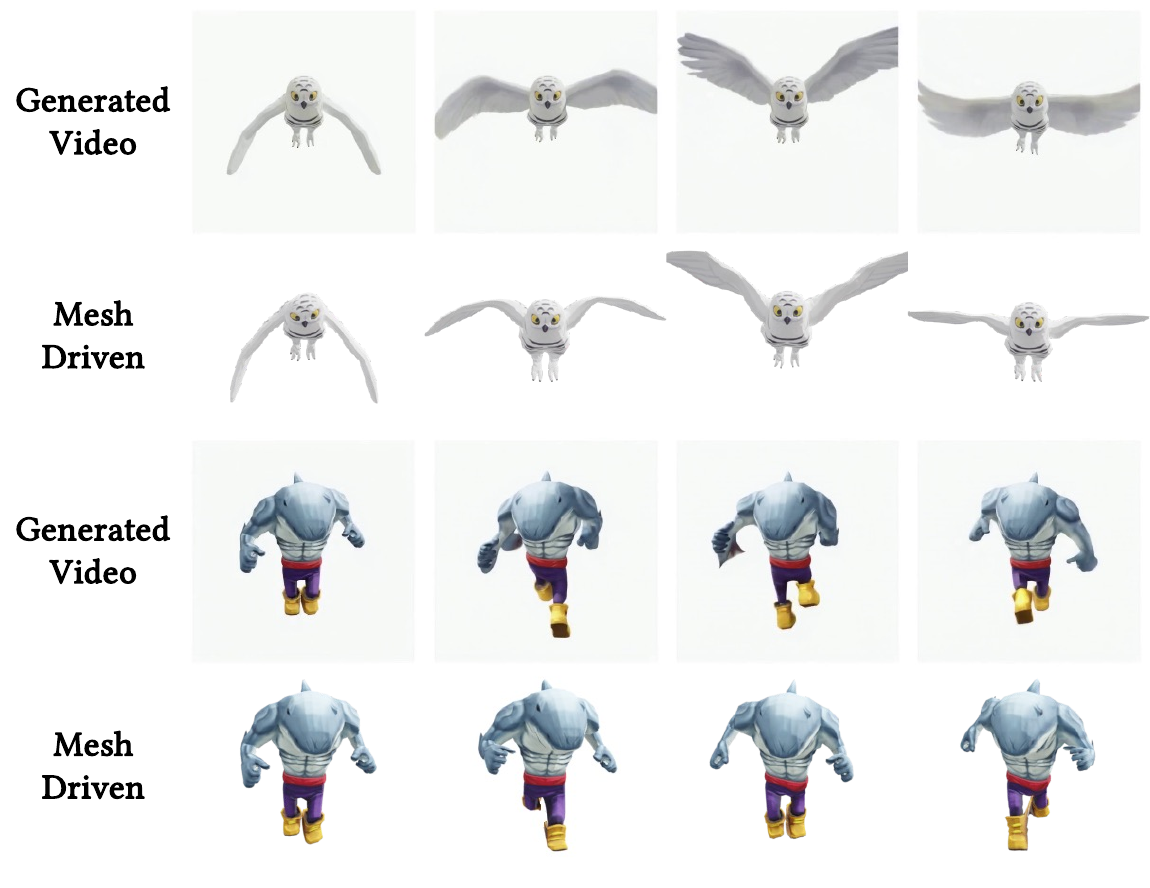}
        \vspace{-12pt}
        \caption{Method usage.}
        \label{fig:sub1}
    \end{subfigure}%
    \hfill
    \begin{subfigure}[t]{0.48\linewidth}
        \centering
        \includegraphics[width=\linewidth]{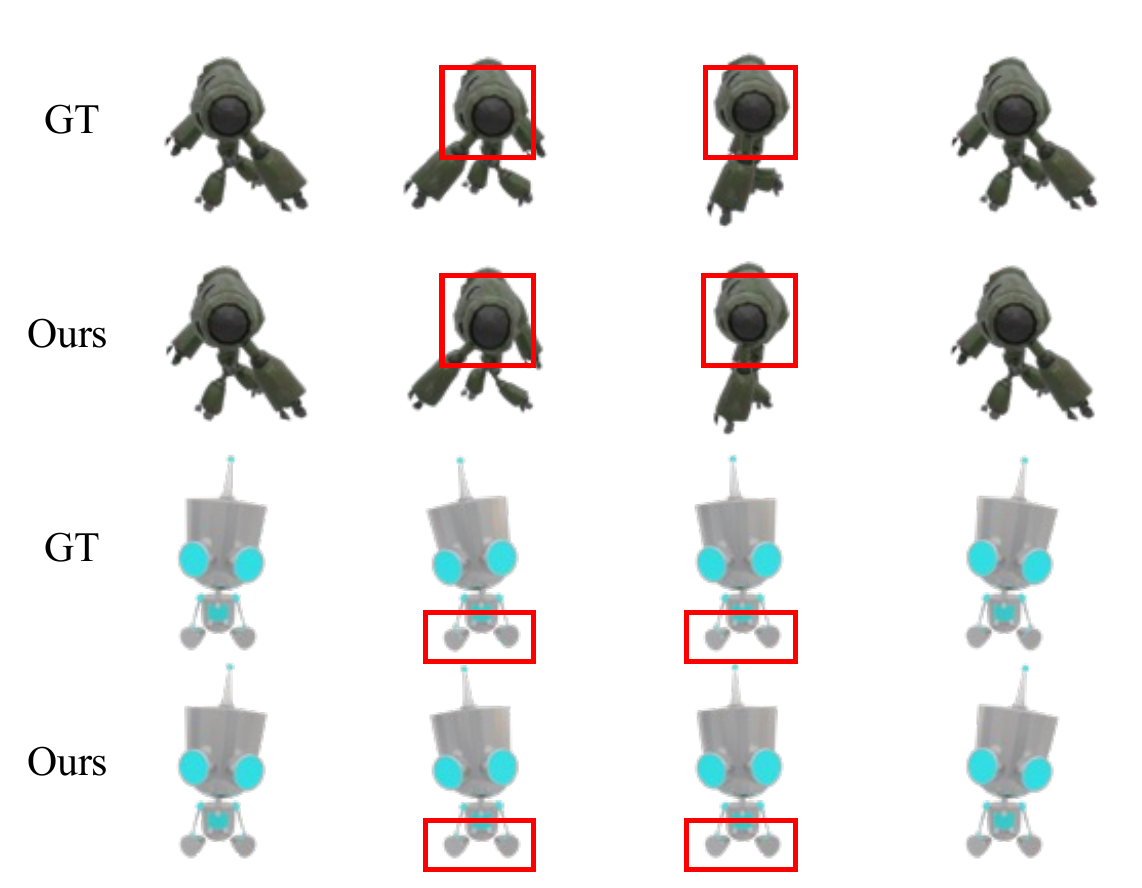}
        \vspace{-12pt}
        \caption{Failure cases.}
        \label{fig:sub2}
    \end{subfigure}
    \vspace{-6pt}
    \caption{Visualization of application and failure cases.}
    \label{fig:both}
\end{figure}

\noindent \textbf{User Study.}
We conduct qualitative comparisons between our proposed method and the baseline models~\cite{dreamgaussian4d, consistent4d, stag4d, dreammesh4d} by surveying 15 human evaluators. To assess the quality of both the appearance and the generated motion, we assign two distinct scores for each video: an appearance score and a motion score. We generate five corresponding videos for each reference video using our method and the baseline models. We implement a ranking system to evaluate these generated videos based on the perceived quality of appearance and motion. Specifically, the evaluators rank the videos for both aspects separately, where the best video in each category is assigned 5 points, the second-best 4 points, and so on, with the worst video receiving 1 point. After collecting the rankings for all evaluators, we compute the average score for all generated videos, corresponding to reference videos, to obtain an overall quality ranking for appearance and motion. The resulting appearance and motion scores are summarized in Tab.~\ref{tab:1}. These subjective evaluations align with the quantitative results, reinforcing the conclusion that, under human evaluation, our method consistently outperforms the baseline models. This indicates that our approach yields superior results in both visual quality and motion generation, not only in objective metrics but also in subjective assessment.

\noindent \textbf{Ablation Study.}
During the Variational Autoencoder (VAE) training, we observed that using only the Mean Squared Error (MSE) loss did not allow for fast convergence to a satisfactory result. We introduced the Euclidean distance loss (DIS) to help optimize the error between the predicted point cloud and the ground truth point cloud. Consequently, we conducted an ablation study and found that, within the same optimization time, the combination of both loss functions led to better results. The PSNR and Chamfer Distance metric is reported in Tab.~\ref{tab:2}. Therefore, we ultimately adopted the combination of both loss functions to optimize the VAE. We also conduct an ablation study on the latent size of the VAE encoder to guide the selection of appropriate hyperparameters in Tab.~\ref{tab:3}.

\noindent \textbf{Method usage.}
We aim to propose a novel approach for generating 4D assets. Our experiments demonstrate that our method can produce a set of point cloud trajectories to drive an initial mesh in a modern rendering engine, achieving improved performance on the proposed dataset. However, aligning 3D assets with corresponding video data is often challenging. Our method has two possible applications: one is to generate a 3D asset from the first frame of a video, and the other is to render an image from a 3D asset, then use video diffusion~\cite{svd} to generate a driving video. To illustrate this, we show our attempts in Fig.~\ref{fig:sub1}, where the generated video drives an existing 3D model, demonstrating the potential applications of our approach.

\begin{table}
\centering
{
\caption{MSE/DIS loss ablation.}
\label{tab:2}
\fontsize{9pt}{10.8pt}\selectfont
\begin{tabular}{c|cc|cc}
\toprule
Case &Using MSE & Using DIS & PSNR$\uparrow$ & CD$\uparrow$  \\
\midrule
\#1 & \ding{51} & - & 23.131 & 0.030 \\
\#2 & - & \ding{51} & 23.739 & 0.023 \\
\#3 & \ding{51} & \ding{51} & \textbf{24.046} & \textbf{0.019} \\
\bottomrule
\end{tabular}
}
\vspace{-12pt}
\end{table}

\begin{table}
\label{tab:3}
\centering
{
\caption{Latent size ablation.}
\label{tab:3}
\fontsize{9pt}{10.8pt}\selectfont
\begin{tabular}{c|cc|cc}
\toprule
Case & $C=128$ & $C=512$ & PSNR$\uparrow$ & CD$\uparrow$  \\
\midrule
1 & \ding{51}($C_0=8$) & - & 22.366 & 0.039 \\
2 & \ding{51}($C_0=16$) & - & 22.897 & 0.031 \\
3 & \ding{51}($C_0=32$) & - & 23.417 & 0.025 \\
4 & - & \ding{51}($C_0=8$) & 23.335 & 0.026 \\
5 & - & \ding{51}($C_0=16$) & 23.852 & 0.021 \\
6 & - & \ding{51}($C_0=32$) & \textbf{24.046} & \textbf{0.019} \\
\bottomrule
\end{tabular}
}
\end{table}

\section{Conclusion}
\label{sec:conclusion}
In conclusion, this paper presents a novel pipeline for generating 4D assets by driving a 3D mesh using point trajectories. Additionally, a corresponding dataset is introduced. To address this challenge, we propose \textbf{DriveAnyMesh}, a method for 4D generation based on 4D latent diffusion, where we employ latent sets to encode both geometry and motion. Through extensive experimentation, our approach demonstrates superior performance compared to existing methods, with experimental results highlighting its effectiveness in practical applications. Our method is constrained by resolution limitations, which may hinder its ability to capture fine-grained motion details. While it successfully generates meshes that preserve topological consistency, the approach operates on a per-point basis, which can lead to challenges in maintaining the shape of rigid objects, as illustrated in Fig.~\ref{fig:sub2}. Future work could explore increasing the training resolution and incorporating rigid relationships between points to address these issues.

{
\small
\bibliographystyle{plain}
\bibliography{ref.bib}
}

\end{document}